\documentclass{article}
\usepackage{spconf,amsmath,graphicx,subfig,multirow,amssymb,pifont}
\usepackage[colorlinks,linkcolor=red]{hyperref}

\title{Embedded Feature Similarity Optimization with Specific Parameter Initialization for 2D/3D Medical Image Registration}
\name{Minheng Chen$^{1}$ \qquad Zhirun Zhang$^{2}$ \qquad Shuheng Gu$^{1}$\qquad Youyong Kong$^{1 \star}$\thanks{This work has been submitted to the IEEE for possible publication. Copyright may be transferred without notice, after which this version may no longer be accessible.}} 

\address{$^{1}$ School of Computer Science and Engineering, Southeast University, China \\
$^{2}$  School of Artificial Intelligence, Southeast University, China \\
$^{\star}$
Jiangsu Provincial Joint International Research Laboratory of Medical Information Processing,\\
School of Computer Science and Engineering, Southeast University, China}

\begin{document}
%
\maketitle
\begin{abstract}
 We present a novel deep learning-based framework: Embedded Feature Similarity Optimization with Specific Parameter Initialization (SOPI) for 2D/3D medical image registration which is a most challenging problem due to the difficulty such as dimensional mismatch, heavy computation load and lack of golden evaluation standard. The framework we design includes a parameter specification module to efficiently choose initialization pose parameter and a fine-registration module to align images. The proposed framework takes extracting multi-scale features into consideration using a novel composite connection encoder with special training techniques. 
 We compare the method with both learning-based methods and optimization-based methods on a in-house CT/X-ray dataset as well as simulated data to further evaluate performance. Our experiments demonstrate that the method in this paper has improved the registration performance, and thereby outperforms the existing methods in terms of accuracy and running time. We also show the potential of the proposed method as an initial pose estimator. The code is available at \href{https://github.com/m1nhengChen/SOPI}{https://github.com/m1nhengChen/SOPI} 
\end{abstract}
\section{Introduction}
\label{sec:intro}

 The goal of image registration is to align images into a common coordinate space from which further medical image analysis, such as image-guided intervention, image fusion for treatment decision, and atlas-based segmentation, can be performed. Medical image registration techniques have progressed rapidly in recent decades, particularly the optimization-based methods, which commonly employs similarity measures such as intensity difference, correlation methods for intra-modality registration and information-theoretic metrics for inter-modality registration~\cite{MR1,CT1,MR2}.

Image registration plays a crucial role in the field of medical image analysis and diagnosis. Despite the fact that there are many empirical techniques, 2D/3D registration is one of the most challenging problems in this field since analytical solutions are probably insufficient given the complexity, variability, and high-dimensionality of typical 2D/3D registration problems. This technique is primarily used for X-ray-based image-guided interventions and surgical image-based navigation, to estimate the spatial relationship between 3D pre-operative CT and 2D intra-operative X-ray. 

In existing techniques, optimization-based 2D/3D registration methods are typically used to achieve accurate 2D/3D registration. In these traditional non-learning-based method~\cite{ncc,NGI,gc}, the 3D X-ray attenuation map is used to create the Digital Reconstructed Radiograph (DRR), which simulates the attenuation of virtual X-rays. An optimizer is used to maximize the similarity between the DRR and X-ray images based on intensity. Although these methods have a high degree of accuracy, they still suffer from some limitations such as small capture range and lengthy computation time. Besides, without proper initialization, using these optimization-based similarity measures~\cite{similarity} can result in the optimization entering local minima and producing inaccurate registration outcomes because they are frequently non-convex and inaccurately represent pose offsets when the perturbations are large.

Recently, with the theoretical development of machine learning techniques, learning-based methods have demonstrated promising quality and speed in a variety of medical image registration tasks~\cite{DP1,DP2,DP3}. Numerous researchers are also attempting to apply theoretical machine learning approaches to 2D/3D registration~\cite{DP4,DP5}. 
For instance, Miao et al.~\cite{shun,miao} proposed a CNN regression approach to align 3D models of a trans-esophageal probe, a hand implant, and a knee implant with 2D X-ray images. Their approach is effective, but obtaining a sufficient number of real annotated x-ray images for training is a significant challenge. Gao et al.~\cite{prost} proposed an end-to-end neural network implemented through a novel Projective Spatial Transformers module(ProST) that they have designed. ProST supports differentiable volume rendering, allowing  the implementation of end-to-end 2D/3D registration model based on deep learning. Houtte et al.~\cite{prostm,prostn} designed a more intricate and unrestricted orthogonal projection network based on Gao's work, expanding its usefulness in clinical settings.
However, the existing methods still have certain limitations.  
Due to the limitations of current imaging geometry and pose variability, there is a lack of a proven method for initializing pose parameters.
More importantly, learning-based methods need a large amount of paired CTs and X-rays. But for certain interventional scenarios such as lumbar spine surgery, it is difficult to collect a large amount of X-ray data for the training of the model. What is more, existing 2D/3D registration methods lack a robust and automatic initialization strategy~\cite{review}.

In this paper, we propose a novel two-stage 2D/3D registration framework, Embedded Feature Similarity Optimization with Specific Parameter Initialization (SOPI), which can align the images automatically without a large amount of real X-ray data for training and weaken the effect of incorrect initialization on the registration algorithm. 
In the framework, we propose a regressive parameter-specific module, Rigid Transformation Parameter Initialization (RTPI) module, to initialize pose parameter and an iterative fine-registration network to align the two images precisely by using embedded features. 
The framework estimates the transformation parameter that best aligns two images using one intra-operative X-ray and one pre-operative CT as input. 
Besides, We introduce a composite connection learning algorithm to capture richer long-range dependencies, extract multi-scale features and build better visual correlations. Finally, a loss using projection mask is proposed to apply to the double backward mechanism~\cite{prost}.

\section{Method}
\subsection{Overview}
\begin{figure}[htbp]
\centering
  \includegraphics[width=0.45\textwidth,height=0.15\textwidth]{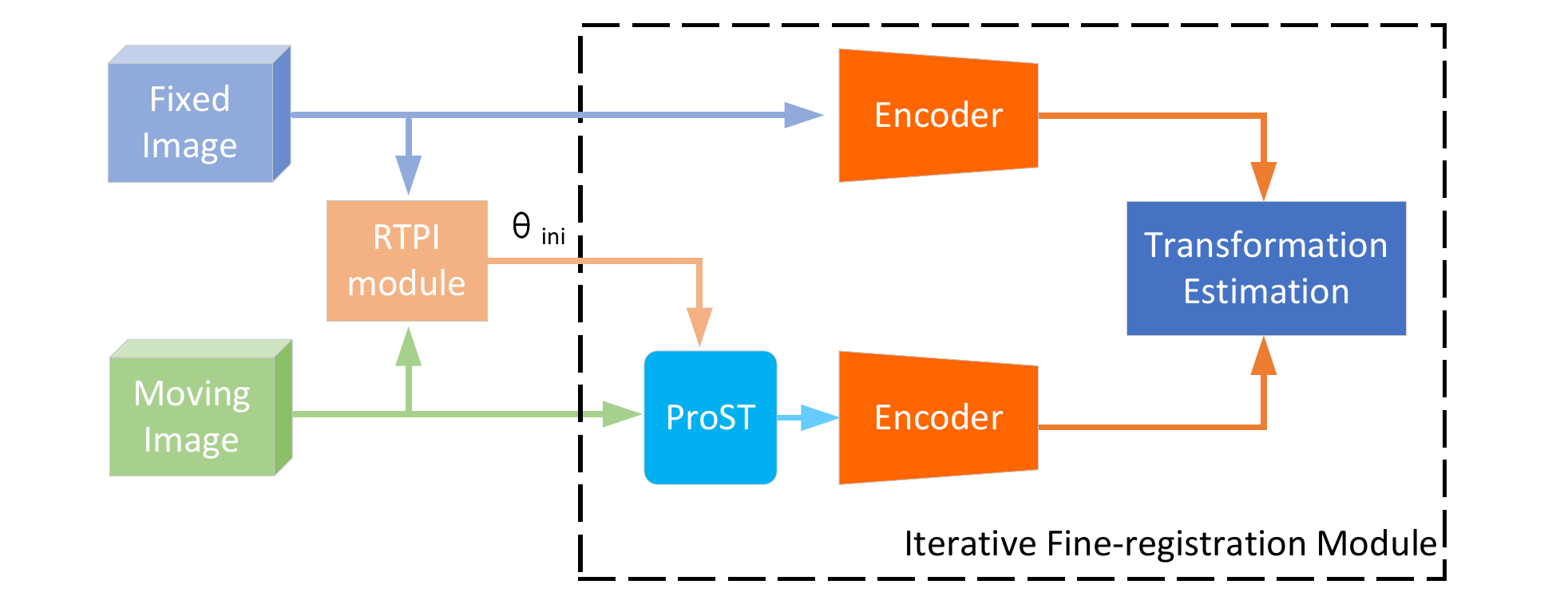} 
\caption{The overall architecture of our proposed registration framework. }
\label{fig:overall}
\end{figure}

Given a fixed 2D X-ray image $\textit{I}_f$  and a moving 3D CT volume $\textit{V}$ as input, the 2D X-ray to 3D CT volume registration problem is to seek a mapping function $\mathcal{F}$ to retrieve the pose parameter $\theta$ such that the image simulated from the 3D CT is as similar as possible to the acquired image $\textit{I}_f$: 
\begin{equation}\label{eq1}
\mathcal{F}(\theta)=arg\mathop{min}_{\theta}\textit{L}_{sim}(\textit{I}_f,\textit{P}(\theta;\textit{V}))
\end{equation}
where $\textit{P}(\theta,\textit{V})$ denotes the mapping from volumetric 3D scene $\textit{V}$ to projective transmission image by using pose parameter $\theta$ and the sampling plane \textit{P} which is generated by projective spatial transformers. 
$\textit{L}_{sim}$ is the similarity function. The parameter $\theta$ contains six degrees of freedom, that is, three in rotation $(\textit{r}_x, \textit{r}_y, \textit{r}_z)$ and three in translation $(\textit{t}_x, \textit{t}_y, \textit{t}_z)$. In this framework, the output of the parameter initialization module $\theta_{ini}$, is used to initialize the parameter in the fine-registration module. And in the iterative fine-registration module, the network will update the pose parameter iteratively until it converges.

\subsection{Rigid Transformation Parameter Initialization Module}

\begin{figure*}[ht]
\centering

\includegraphics[width=\textwidth]{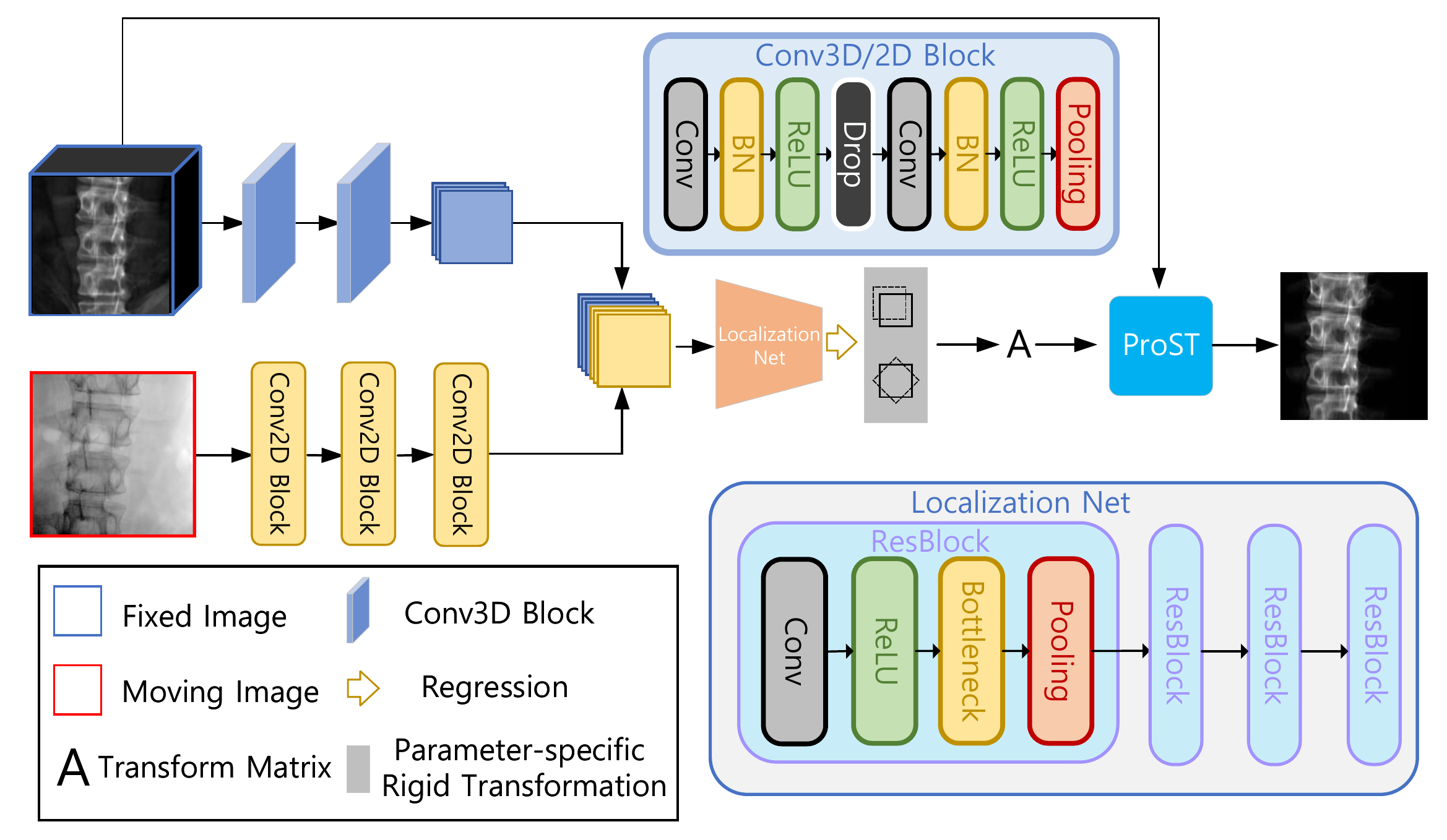} 
\caption{The architecture of the proposed Rigid Transformation Parameter Initialization (RTPI) module. An asymmetric dual-branch structure has been created to extract features independently in order to balance data information. And a parameter-specific method has been used to regress each geometric transformation parameter respectively.  }
\label{fig:RTPI module}
\end{figure*}

The input to our network is a CT volume \textit{V} and an X-ray image $\textit{I}_f$. 
The proposed parameter initialization Module is shown in Fig.~\ref{fig:RTPI module}. The dimension gap between 2D and 3D images is a significant barrier to the performance of registration. Directly concatenating these two inputs together would make the network overlook the contents of the 2D image due to the information asymmetry. We create an asymmetric dual-branch structure to extract features independently in order to balance data information. In the 3D image branch, we use two 3D convolution blocks to extract the volume feature, and then flatten the output feature map to 2D. Through this approach, the flattened feature maps of the two branches are concatenated along the channel dimension, which makes them have identical size. Through this, it achieves the desired information balance. The localization net we use is implemented by four residual blocks~\cite{resnet}. 

The rigid transformation matrix \textbf{A} can be formed by composing the rotation matrix $\textbf{M}_r$ and the translation matrix $\textbf{M}_t$ in turn,
\begin{equation}\label{matrix}
\textbf{A}=\textbf{M}_t \cdot \textbf{M}_{rx} \cdot \textbf{M}_{ry} \cdot \textbf{M}_{rz}=\begin{bmatrix} a_1 & a_2 & a_3 & a_4 \\ a_5 & a_6 & a_7 & a_8\\ a_9 & a_{10} & a_{11} & a_{12}\\ 0 & 0 & 0 & 1\end{bmatrix}
\end{equation}
and each of the four transformation matrix can be represented as: 
\begin{equation}\label{matrix_1}
\begin{aligned}
\textbf{M}_t&= \begin{bmatrix} 1 & 0 & 0 & t_x \\ 0 & 1 & 0 & t_y\\ 0 & 0 & 1 & t_z\\ 0 & 0 & 0 & 1\end{bmatrix} \\
\textbf{M}_{rx}&=\begin{bmatrix} 1 & 0 & 0 & 0\\0 & cos(r_x) & -sin(r_x) & 0 \\ 0 & sin(r_x) & cos(r_x) & 0\\  0 & 0 & 0 & 1\end{bmatrix} \\
\textbf{M}_{ry}&=\begin{bmatrix} cos(r_y) & 0 & sin(r_y) & 0 \\ 0 & 1 & 0 & 0\\ -sin(r_y) & 0 & cos(r_y)& 0\\ 0 & 0 & 0 & 1\end{bmatrix} \\
\textbf{M}_{rz}&=\begin{bmatrix} cos(r_z) & -sin(r_z) & 0 & 0 \\  sin(r_z) & cos(r_z)& 0 & 0\\ 0 & 0 & 1 & 0\\ 0 & 0 & 0 & 1\end{bmatrix} \\
\end{aligned}
\end{equation}
To ensure that these spatial transformation parameters are unique solutions and to enhance the interpretability of the effects of each type of transformation, we use a parameter-specific method~\cite{PASTA} to regress each geometric transformation parameter instead of directly regressing the transformation matrix \textbf{A} before using Projective Spatial Transformer module (ProST)~\cite{prost}. Furthermore, this method could effectively prevent the problem of generating out-of-plane images by the regression parameters and break constraint of current imaging geometry. At the end of the network, a projective image $\textit{I}_m$ is generated by ProSTs from CT volume and transformation matrix \textbf{A}.

\begin{figure*}
    
\centering
\includegraphics[width=\textwidth,height=0.45\textwidth]{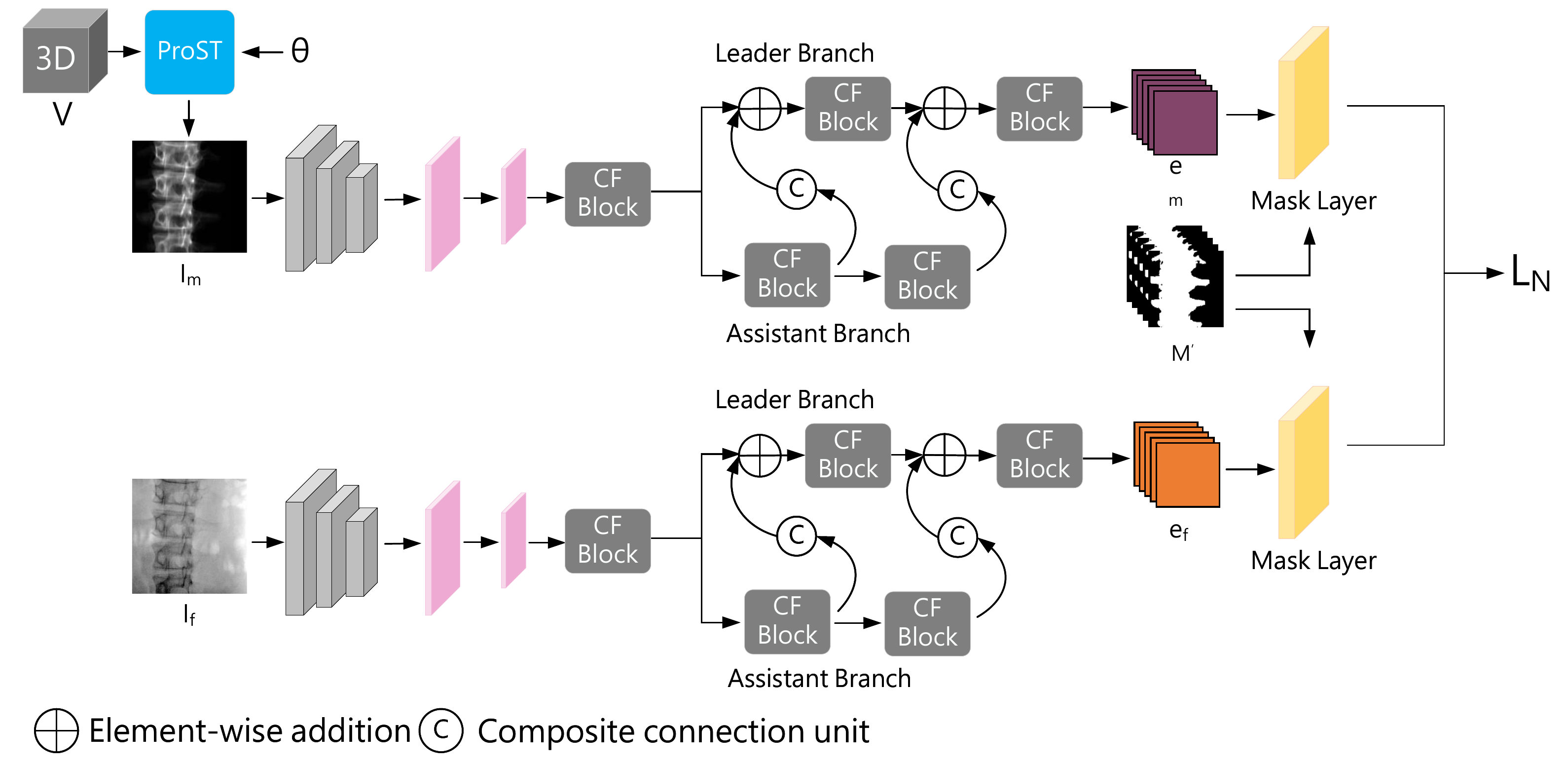} 
\caption{The architecture of the proposed iterative fine-registration module. This module mainly consists of two proposed composite encoders. The encoder consists of two branches, the assistant branch extracting the low-level feature which will be fused to the leader branch by composite connection unit and concat operation.So on the leader branch we can get a combination of high-level and low-level features which could better represent the  correlation between images}
\label{fig:fine-registration}
\end{figure*}

In many unsupervised image registration methods, 
similarity metrics for conventional optimization-based registration (e.g. NCC, MI, GC)~\cite{similarity} are successfully utilized to be the loss functions for gradient-descent optimization. However, the drawback of the similarity metric is that it is easy to get trapped in local minima. Thus a supervised mean squared error (MSE) loss can be computed, which directly uses ground-truth labeling information $\hat{\theta}$ to constrain the $\theta$ estimation:\begin{equation}\label{mse}
\textit{L}_\textit{mse}(\hat{\theta}, \theta) = \cfrac{1}{\textit{N}}\sum^{\textit{N}}_{i=1}\Vert \hat{\theta}_{i}-\theta_{i} \Vert_2
\end{equation}
Through the mean square error and similarity loss mentioned above, we update rigid transformation parameter initialization network using the following loss function:\begin{equation}\label{rtpi}
\textit{L}_\textit{RTPI module} = \alpha\textit{L}_\textit{sim}(\textit{I}_f,\textit{I}_m) + \beta\textit{L}_\textit{mse}(\hat{\theta}, \theta) + \lambda\mathcal{R}(\theta)
\end{equation}
where $\textit{L}_\textit{sim}$ is the gradient difference loss ~\cite{comparison}.
We consider using other similarity metrics, e.g. normalized cross correlation, mutual information, normalized gradient correlation, as $\textit{L}_\textit{sim}$ in Eq.\ref{rtpi} for training, but we find that the network could not converge.  
$\mathcal{R}(\theta)$ is a L2 regularization term with the regularization parameter $\lambda$. In addition, $\alpha$, $\beta$ are hyper-parameters.
\subsection{Iterative Fine-Registration Module}
Fig.~\ref{fig:fine-registration} shows the module architecture. The input includes a 3D volume and a pose
parameter $\theta$ and a fixed image: $\textit{I}_f$. Projection operation is performed by ProST~\cite{prost} with respect to $\theta$. The projected moving image $\textit{I}_m$ and the input fixed image $\textit{I}_f$ go through two encoders, which are the same in structure but do not share the weights, and output embedded features $\textit{e}_m$ and $\textit{e}_f$.

This module is designed for fine registration, so it is critical to design a network architecture for multi-scale feature fusion that could extract the fine-grained texture features of image while paying attention to high-level features, which could expand the capture range.

We absorb the advantages of two network architecture,  Cascade fusion network~\cite{cfnet} and composite backbone network~\cite{cbnet1}, to propose an composite encoder(CE) for multi-scale feature fusion that extracts both semantic and global information into embedded features. The encoder consists of two branches, the assistant branch which extracts the low-level feature  will be fused to the lead branch by composite connection unit and concat operation. Therefore, on the leader branch, we can obtain high-level and low-level features that can better represent the correlation between pictures. Each composite connection unit consists of both a 1×1 convolutional layer and batch normalization layer to reduce the channels, and an upsample operation. Before splitting encoder into two branches, we first use a three layer convolution network to extract shallow features from the image, after each layer of convolution, there is an activation function and a normalization layer. Next, we use two convolutional layers to downsample the features followed by a CF block.

There are two cascaded fusion (CF) blocks on each branch of the encoder. And the bottleneck in each cascade fusion block is implemented by using split-attention~\cite{resnest}. Let the size of the embedded features at the output of the encoder be $H \times W \times C$. We define a mask \textit{M} of the same size as the input 3D volume to demarcate the voxels belonging to the spine and use $M^{'}=\textit{P}(\theta;M)$ to make sure the loss is applied only to a portion of the online-sampled projected image in order to reduce computational effort without losing semantic information . Then we downsample $M^{'}$ to $H \times W$ and stack it to the size same as embedded features. In mask layer, we extract the features which are equal to one in $M^{'}$ from the embedded feature map.

\begin{figure*}[ht]

\centering
  \includegraphics[width=\textwidth]{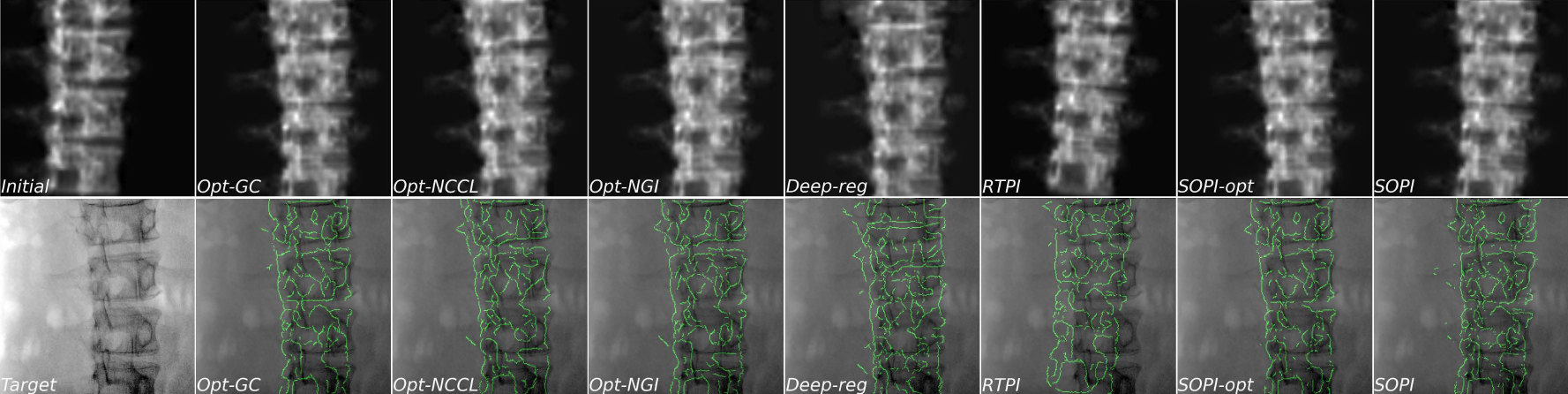}
\caption{Qualitative examples of our method and the baseline methods. The first row shows the projection results of the postures predicted by each method, and the second row shows the fusion images with the X-ray image, respectively.}
\label{fig:fusion_map}
\end{figure*}

The training strategy of this module is based on geodesic gradient
which has been studied for registration problems due to its convexity~\cite{gg1,gg2}, 
and double backward mechanism which means computing parameter gradient by using a error function in first back-propagation and updating the network parameters by the real loss function $L$ during second back-propagation. The loss function of our network is to approximate the gradient of the network back-propagating from error function $L_N$  to $\theta$ and the geodesic gradient $L_G$ between $\theta$ and the pose of the fixed image $\hat{\theta}$. So the network error function can be formulated as:
\begin{equation}\label{error function}
L_N (\textit{e}_m,\textit{e}_f)= \frac{\sum_i^{H\times W \times C}M_i^{'}\cdot \Vert \textit{e}_{m_i}-\textit{e}_{f_i} \Vert_2}{\sum_i^{H\times W \times C}M_i^{'}}
\end{equation}
During the process of back-propagation, we divide the gradient vector into two components, rotation vector $V_r$ and translation vector $V_t$. We define the gradient of the network back-propagating to $\theta$ as $\frac{\partial L_N}{\partial \theta}=\{V_r,V_t\}$  and the pose geodesic gradient as $\frac{\partial L_G(\theta,\hat{\theta})}{\partial \theta}=\{V_r^{'},V_t^{'}\}$. Thus the final loss function is
\begin{equation}\label{fine loss}
L= \Vert\frac{V_r^{'}}{\Vert V_r^{'}\Vert_2}- \frac{V_r}{\Vert V_r\Vert_2}\Vert_2 +\Vert\frac{V_t^{'}}{\Vert V_t^{'}\Vert_2}- \frac{V_t}{\Vert V_t\Vert_2}\Vert_2
\end{equation}

At inference phase, we fix the network parameters and start with an initial pose $\theta$ given by RTPI module. We can perform gradient-based optimization over $\theta$ based on the back-propagation gradient flow. 
Instead of optimizing the rotation and transition components at the same time consistently, we fix the rotational component after a few iterations and continue iterating until convergence due to the fact that the two components do not converge at the same rate. 
The network similarity is more effective compared to traditional optimization-based methods, allowing faster convergence and high accuracy due to the structure of the encoder which help to extract rich multi-scale features with few float-point operations.

\section{Experiment}

\subsection{Datasets and Experimental Setting}
The dataset we use in the experiments contains 146 raw CT scans collected  from several partner hospitals. The spine is segmented using an automatic method in~\cite{segmentation}. Segmentations and its mask are cropped to the lumbar spine cubic region and downsampled to the size of $128 \times 128 \times 128$. We use 136 scans for training, and 10 scans for validation. Furthermore, we evaluate the performance of our method by using a set of CTs and corresponding lumbar fluoroscopy from 5 acquisitions, in addition to the simulations generated from the validation set. We define the intrinsic parameter of the ProST module following a Perlove PLX118F C-Arm, which generates the X-ray image for our experiments. We set isotropic pixel spacing to 0.19959 mm/pixel, the source-to-detector distance to 1011.7 mm and we downsample the detector dimension from $1024 \times 1024$ to $256 \times 256$. The implementation is based on PyTorch. The experiments are performed on a workstation equipped with four NVDIA RTX 3090 GPUs and four 8-core Intel CPUs. 
\subsection{Training Details}
Our framework is trained in two steps using the same mechanism. The  RTPI module is trained on simulated data generated by ProST. At training iteration i, we randomly sample a pair of pose parameters, ($\theta^i$, $\theta^i_t$), rotation from $N$(0, 20) in degree, in all three axes, $t_x,t_y,t_z$ from $N$(0, 100), $N$(0, 30), $N$(0, 15) respectively in mm. We then randomly select a CT segmentation $V_{seg}$. The target image is generated online from $V_{seg}$ and $\theta^i_t$ using ProST. $V_{seg}$ and $\theta^i$ are used as input to our network forward pass. The network is trained using SGD optimizer with a cyclic learning rate between 1e-3 and 1e-2 every 100 steps and a momentum of 0.9. Batch size is chosen as 16 and we train 200k iterations until convergence. For the parameters in Eq.\ref{rtpi},  we set $\alpha=1$, $\beta=1$, and $\lambda= 0.01$. For implementation of Eq.\ref{fine loss},We use the geodesic gradients of $\theta^i$ and $\theta^i_t$ in the Riemannian metric, which are implemented by~\cite{geometry}. 

The iterative fine-registration module is trained using the same mechanism as the RTPI module, but with different setting of pose parameter. For the paired pose parameters, ($\theta^i$, $\theta^i_t$), we randomly sample rotation from $N$(0, 20) in degree, translation from $N$(0, 20) in mm. We then randomly select a CT segmentation $V_{seg}$ and its mask $M$. The fixed image $I_f$ and $M^{'}$ is generated online from $\textit{P}(\theta^i_t;V_{seg})$ and $\textit{P}(\theta^i;M)$ respectively. $V_{seg}$, $ M^{'}$ and $\theta^i$ are used as input to our network forward pass. We use the geodesic gradients of $\theta^i$ and $\theta^i_t$ in the Riemannian metric, which are implemented by~\cite{geometry}.
The network is trained for 200k iterations with batch size $K=4$ using SGD optimizer with a cyclic learning rate between 1e-4 and 1e-3 every 100 steps and a momentum of 0.9.

\subsection{Evaluation and Results}
\label{sec:eval}

We compare our method (SOPI) with one learning-based and four optimization-based methods. To further evaluate the performance of the proposed method as an initial pose estimator, we also demonstrate the performance of the method using our SOPI to initialize the optimization on X-ray data. We denote this approach as SOPI+opt.  
For SOPI+opt, we use the Opt-GC method during the optimization as we find it converges faster when the initial pose is close to the global optima.

\noindent\textbf{Simulation Study.} 
we analyze the performance of the proposed approach based on simulation studies. We performed the 2D/3D registration application by randomly choosing a pose pair from the same training distribution ($\theta^i$, $\theta^i_t$) of RTPI module. 
Besides the mean translation and rotation error, we also determine a registration failure rate as the percentage of the unsuccessful tested cases. A registration is considered to be successful if the remaining translation and/or rotation error is smaller than 3 mm or 3$^{\circ}$, respectively. In addition, average registration time is reported. The results are shown in Table.~\ref{simulation}. And we find that our method outperforms the baseline methods on every metric.
\begin{table}[h!]
\footnotesize
\begin{center}
	\label{table3}
 
	\caption{performance comparison between our method and baseline methods on simulation.}
	\begin{tabular}{l|l|l|l|l} 
		\hline
  \multirow{2}{*}{Method} &  \multirow{2}{*}{Rotation($^{\circ}$)} & \multirow{2}{*}{Translation(mm)} & Failure& Reg.\\
        &  & &rate(\%) & time\\
		\hline
  Initial
&6.40$\pm$3.77&15.08$\pm$8.56&95.2&N/A\\
            
Opt-NCCL~\cite{ncc}
&3.68$\pm$3.18&5.79$\pm$5.18&38.2&19.16 \\

Opt-NGI~\cite{NGI}
&3.84$\pm$3.32&5.92$\pm$5.40&50.6&30.25 \\

Opt-GC~\cite{gc}
&3.73$\pm$3.18&7.80$\pm$7.25&43.6&18.74 \\

Deep-reg~\cite{prost}
&5.54$\pm$3.83&13.21$\pm$8.55&74.0&20.09 \\
            
SOPI
&\textbf{1.89$\pm$1.57}&\textbf{4.53$\pm$3.54}&\textbf{22.4}&\textbf{4.64} \\
            \hline
	\end{tabular}
\label{simulation}	
\end{center}
\end{table}

\noindent\textbf{Real X-ray Study.} 
For real X-ray data, the performances of the proposed method and the baseline methods are evaluated with distance error(DistErr), i.e. the mean distance (in mm) between the ground truth X-ray and the predicted-sampled DRR's corresponding corner points by using~\cite{surf,orb}. The image similarity score(ImgSim) shows the quality assessment of the registration results. We also report the rotation and translation errors in  degree and millimeter respectively. In addition, the registration time denotes the average time cost for registering a single pair of images.
The evaluation results are given in Table.\ref{eval}.

We find that Opt-GC method has the higher accuracy in rotation than the other optimization-based methods. 
By applying our method, a competitive performance has been achieved compared to the existing methods. Our running time is nearly ten times faster than optimization-based methods, and compared to Deep-reg, we have higher accuracy and less computation time. In addition, we notice that when our method is combined with an optimization-based method, the registration error is greatly reduced and the registration speed is also significantly improved, which demonstrates that our module offers initial pose much closer to the global optima. Fig.\ref{fig:fusion_map} shows the qualitative examples of our method and the baseline methods.

\noindent\textbf{Ablation Study.}
We conduct an additive ablation study on the simulated data to illustrate whether the superior performance of our method is attributed to the initial parameter provided by RTPI module, the effectiveness of the proposed composite encoder structure and the embedded feature loss using mask projection.
\begin{table}[htbp]
\begin{center}
	\label{table2}
 
	\caption{Ablation study results on: 1) Rigid Transformation Parameter Initialization module (RTPI) 2) Composite Encoder (CE) and 3) Embedded-feature Loss (EL)}
	\begin{tabular}{c|c|c|l|l} 
		\hline
  \multirow{2}{*}{RTPI} &\multirow{2}{*}{CE} &\multirow{2}{*}{EL} &  \multirow{2}{*}{Rotation($^{\circ}$)} & \multirow{2}{*}{Translation(mm)} \\
      & & &  & \\
		\hline



\checkmark&\ding{53} & \checkmark

&5.42$\pm$3.71&12.06$\pm$8.89\\
\ding{53}&\checkmark &\checkmark
&5.67$\pm$3.72&11.00$\pm$8.17\\
\checkmark&\ding{53} & \ding{53}

&3.42$\pm$3.30&6.68$\pm$4.18\\
 \checkmark&  \checkmark& \ding{53}

&2.09$\pm$2.46&4.53$\pm$3.92\\
 \checkmark& \checkmark& \checkmark
&\textbf{1.89$\pm$1.57}&\textbf{4.41$\pm$3.54} \\
            \hline
	\end{tabular}
\label{ablation}	
\end{center}
\end{table}
As shown in Table.\ref{ablation}, 
we use mean rotation and translation error to evaluate different models with specific design choices.
after removing RTPI modules, the composite encoder and $L_N$  respectively, the performance of the model decline to a certain extent, which implies that all the network modules in our method are effective for 2D/3D registration. Using RTPI module alone can achieve a certain registration accuracy with very little running time, however, we do not recommend it due to the poor robustness.


\begin{table*}[htbp]

\begin{center}
	\label{table1}
 
	\caption{2D/3D registration performance comparing with the baseline methods on X-ray. The performances of the proposed method and the baseline methods are evaluated with distance error(DistErr), the image similarity score(ImgSim) and  average registration time. Additionally, we also report the rotation and translation errors in degree and millimeter respectively. }
	\begin{tabular}{l|l|l|l|l|l|l|l|l|l} 
		\hline
  \multirow{2}{*}{Method} &DisErr & ImgSim & \multicolumn{3}{l|}{Rotation($^{\circ}$)} & \multicolumn{3}{l|}{Translation(mm)} & Reg.\\
  \cline{4-9}
		& (mm) &(NCC) & rx & ry & rz & tx & ty & tz & time\\
		\hline
Initial &0.036&0.462&8.66&5.01&4.52&13.41&16.13&16.26& N/A\\
Opt-NCCL\cite{ncc} &0.032&0.962&7.55&2.40&1.15&7.10&4.85&3.10&31.44 \\
Opt-NGI\cite{NGI}
&0.036&0.941 &7.75&5.00&1.30&18.15&4.80&\textbf{2.30}&48.64 \\
Opt-GC\cite{gc}
&0.027&0.939&4.70&2.40&1.60&7.15&3.15&3.00&40.14\\
Deep-reg\cite{prost}
&\textbf{0.017}&\textbf{0.963}&6.80&7.93&4.07&5.82&6.67&10.08&20.08\\

SOPI
&0.020&0.905&\textbf{2.83}&\textbf{0.90}&\textbf{0.88}&\textbf{5.64}&\textbf{1.50}&3.30&\textbf{4.67}\\
            \hline
SOPI+opt
&0.013&0.955&2.1&0.47&0.20&1.93&0.65&0.45 &33.80\\
\hline
	\end{tabular}
\label{eval}	
\end{center}
\end{table*}
\section{Limitation}

\begin{figure}[ht]
\centering
  \includegraphics[width=0.45\textwidth, height=0.2\textwidth]{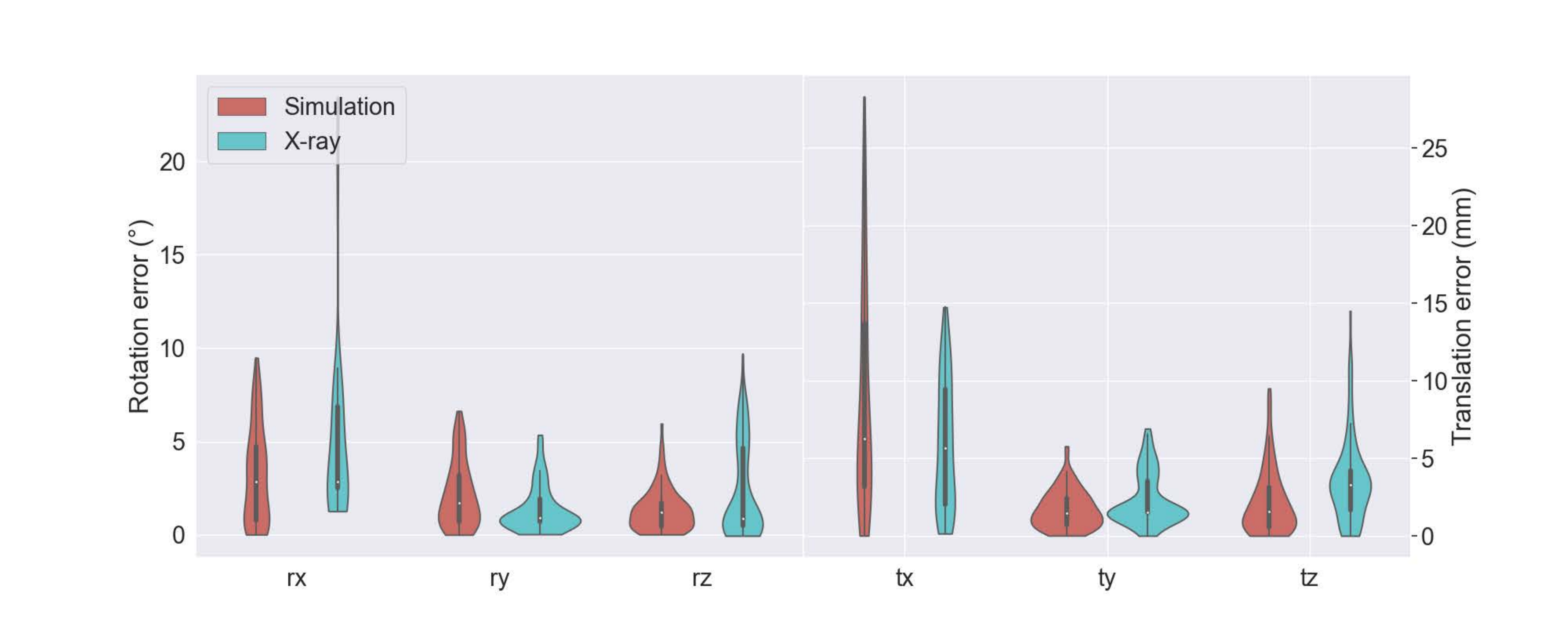}
\caption{Comparison of simulated data and X-ray error distribution}
\label{fig:distribution}
\end{figure}

First,  as shown in Fig.~\ref{fig:distribution}, we also compare the error distribution of our method on simulated data and real X-ray data. The result demonstrates that training our method on the simulated data generated by projective spatial transformers, which has the advantage of learning reliable feature representations without a sizable dataset, will also leads to some degradation in its performance on real data due to the discrepancy between simulated data and x-rays. 

Second, Although our two-stage architecture is fast and robust and can achieve high accuracy when combined with optimization-based methods, it is still desirable to come up with more elegant solutions to solve the problem directly.

\section{Conclusion }
In this paper, we proposed a fast and robust method for 2D/3D registration. 
The proposed method provides a parameter initialization method by directly using the RTPI module, and a fine-registration module that uses the same iterative optimization design as the optimization-based registration methods,  where the difference is that we took the advantage of the powerful perceptual and expressive capabilities of deep networks to learn a set of more complicated features than the conventional hand-crafted ones. 
We evaluated the proposed two-stage framework on a challenging clinical scenario,
i.e. orthopedic lumbar surgery navigation
and showed that our method is significantly more robust and faster than the state-of-the-art optimization-based approaches and learning-based approach. We also demonstrated the potentiality of using our SOPI framework as a pose initial estimator to improve the speed of the current optimization-based approach while attaining a higher registration accuracy. 
In addition, we observed the discrepancies in the performance of our method on simulated data and X-rays, which we believed this is due to the limitation of the projection imaging algorithms and will work on that as a future direction.

\bibliographystyle{IEEEbib}
\bibliography{strings,refs}

\end{document}